\documentclass{article}

\PassOptionsToPackage{numbers, compress}{natbib}
    \usepackage[final]{neurips_2020}

\usepackage[utf8]{inputenc} % allow utf-8 input
\usepackage[T1]{fontenc}    % use 8-bit T1 fonts
\usepackage{hyperref}       % hyperlinks
\usepackage{url}            % simple URL typesetting
\usepackage{booktabs}       % professional-quality tables
\usepackage{amsfonts}       % blackboard math symbols
\usepackage{nicefrac}       % compact symbols for 1/2, etc.
\usepackage{microtype}      % microtypography

\usepackage{amsmath}
\usepackage{graphicx}
\usepackage[ruled]{algorithm}
\usepackage[noend]{algpseudocode}
\usepackage{mathtools}

\title{Assessing SATNet's Ability to Solve the \\Symbol Grounding Problem}

\author{%
  Oscar Chang, Lampros Flokas, Hod Lipson\\
  Data Science Institute\\
  Columbia University\\
  \texttt{\{oscar.chang,lampros.flokas,hod.lipson\}@columbia.edu} \\
   \And
   Michael Spranger \\
   Sony AI \\
   \texttt{michael.spranger@sony.com} \\
}

\begin{document}

\maketitle
% - remove flaw/bug reference
% - incorporate discussion from rebuttal about end-to-end and significance
% - fix shit for m-aux exp
% - rewrite abstract/intro
% - broader impact: reiterate why symbol grounding is important
% - ninth page is allowed
% - include new citations from reviewer 4
\begin{abstract}
SATNet is an award-winning MAXSAT solver that can be used to infer logical rules and integrated as a differentiable layer in a deep neural network \citep{wang2019satnet}. It had been shown to solve Sudoku puzzles visually from examples of puzzle digit images, and was heralded as an impressive achievement towards the longstanding AI goal of combining pattern recognition with logical reasoning. In this paper, we clarify SATNet's capabilities by showing that in the absence of intermediate labels that identify individual Sudoku digit images with their logical representations, SATNet completely fails at visual Sudoku (0\% test accuracy). More generally, the failure can be pinpointed to its inability to learn to assign symbols to perceptual phenomena, also known as the symbol grounding problem \citep{harnad1990symbol}, which has long been thought to be a prerequisite for intelligent agents to perform real-world logical reasoning. We propose an MNIST based test as an easy instance of the symbol grounding problem that can serve as a sanity check for differentiable symbolic solvers in general. Naive applications of SATNet on this test lead to performance worse than that of models without logical reasoning capabilities. We report on the causes of SATNet’s failure and how to prevent them.

%This underscores a major caveat to SATNet's claim to integrate logical structures with deep learning: it is possible to pre-train a digit classifier with the labels, and then use SATNet, independently of a deep network, to solve for Sudoku. So doing can even be advantageous, because the presence of intermediate labels does not guarantee that SATNet learns to assign the correct symbols to perceptual phenomena, which is a problem known as symbol grounding \citep{harnad1990symbol} and long thought to be a prerequisite for intelligent agents to perform real-world logical reasoning. We propose an MNIST-based test as an easy instance of the symbol grounding problem, and found that naive applications of SATNet on this test lead to performance worse than that of models without logical reasoning capabilities. We also report on causes of SATNet's learning failures and ways to prevent them.
\end{abstract}

\section{Introduction}
\label{section:introduction}
Machine learning systems have become increasingly capable at a wide range of tasks, with neural network based models outperforming humans at tasks like object recognition \citep{russakovsky2015imagenet}, speech recognition \citep{xiong2017the,amodei2016deep}, the game of Go \citep{silver2016mastering,silver2017mastering}, Atari videogames \citep{badia2020agent57,ecoffet2020first}, and more. Nonetheless, the success of deep learning comes with significant caveats: neural networks require immense amounts of labeled data for training, can be easily tricked by tiny input perturbations or spurious correlations, and succumb to brittle generalization when tested on data that deviate ever so modestly from the training distribution. Critics point to these caveats as evidence that deep learning, in its current incarnation, is really just performing a sophisticated type of pattern matching, the likes of which can only ever constitute intelligence in narrow, circumscribed domains \citep{marcus2020next,chollet2019measure}.

By comparison, human intelligence can be applied more generally. This has been argued to be a result of two distinct modes of cognition: \emph{System~1} and \emph{System~2} \citep{evans1984heuristic,kahneman2011thinking}. System~1 happens quickly and without conscious effort, for example comparing the size of objects or locating the general source of a sound. On the other hand, System~2 involves slow and deliberate attention, for example solving for a complicated arithmetic equation or checking that an argument is logical. Current machine learning systems have been likened to System~1 \citep{bengio2019from}, because System~1 mostly involves the use of associative memory, and is highly susceptible to cognitive biases and sensory illusions. Symbolic AI algorithms that are based on logic and search more closely resemble System~2.

To achieve robust human-level AI that can solve non-trivial cognitive tasks, it is crucial to combine both System~1 like \emph{pattern recognition} and System~2 like \emph{logical reasoning} capabilities in a seamless \emph{end-to-end learning} fashion. This is because in many practical problems of interest, it is difficult and expensive to collect intermediate labels to train specific machine learning sub-components. For example, it appears infeasible to build a `danger' classifier for a self-driving car, where every possible dangerous scenario is pre-determined and categorized beforehand. Researchers are thus far able to combine both capabilities in a single AI system, but not train them end-to-end. Famously, OpenAI's very impressive achievement of controlling a robotic hand to solve a Rubik's cube required the separate use of a machine learning system to perform the dexterous manipulation and a discrete solver to decide the side of the cube that should be turned \citep{akkaya2019solving}.

Attempts to bridge the two capabilities seamlessly belong to one of three approaches. The first involves augmenting deep learning models with soft logic operators
\citep{hu2016harnessing,rocktaschel2017end,cingillioglu2018deeplogic,evans2018learning,serafini2016logic,sourek2015lifted,van2020analyzing} or combinatorial solving modules \citep{vlastelica2019differentiation,rolinek2020deep,tschiatschek2018differentiable,blondel2020fast,amos2017optnet}. However, this approach typically requires the programmer to pre-specify intricate logical structures according to the problem domain. Moreover, these logical components are fixed and not amenable to learning. The second approach uses sub-symbolic reasoning techniques like Recurrent Relation Networks to implicitly pick up on logical structures within the problem \citep{santoro2017simple,palm2018recurrent,selsam2018learning}. This approach improves on the first by learning the logical structure implicitly by optimization, but nevertheless also necessitates careful feature engineering. The third approach is the field of inductive logic programming (ILP), which starts from a traditional symbolic AI model like a knowledge base, and adds learning capabilities to it \citep{manhaeve2018deepproblog,yang2017differentiable,cropper2020turning,cohen2016tensorlog}. Unfortunately, ILP is limited to symbolic inputs and outputs, unlike deep neural networks.

Against the backdrop of such approaches, SATNet \citep{wang2019satnet} promised to integrate ``logical structures within deep learning'' with a differentiable MAXSAT solver that can infer logical rules and be used as a neural network layer. SATNet claimed to have solved problems that were ``impossible for traditional deep learning methods and existing logical learning methods to reliably learn without any prior knowledge,'' most notably solving a Sudoku puzzle visually from images of puzzle digits, and was awarded with a Best Paper Honorable Mention at 2019's \textit{International Conference on Machine Learning}.

Based on SATNet's success, one might think that enabling end-to-end gradient-based optimization (i.e.~making every component in a system differentiable) is sufficient for end-to-end learning (i.e.~learning without intermediate supervision signals). However, defining gradients for an objective does not, on its own, result in successful learning outcomes, as exemplified by the history of deep learning. Successful training of architectures with hundreds of layers, where gradients are trivially well defined, is highly non-trivial and requires careful initialization, batch normalization, adaptive learning rates, etc. Additionally, without an appropriate inductive bias (like the rules of the game), learning to solve complex problems like visual Sudoku from relatively few samples is extraordinarily challenging. It is unlikely that end-to-end gradient-based optimization by itself will, in general, result in models that generalize well. 

Thus, SATNet's claim to have solved the end-to-end learning problem of visual Sudoku ``in a minimally supervised fashion'' should be revisited. \textbf{Can SATNet learn to assign logical variables (symbols) to images of digits (perceptual phenomena) without explicit supervision of this mapping?} This is also known as the symbol grounding problem \citep{harnad1990symbol}, which has long been thought to be a prerequisite for intelligent agents to perform real-world logical reasoning. If answered in the affirmative, SATNet would have marked a revolutionary leap forward for the whole field of AI, by virtue of the difficulty of the symbol grounding problem in visual Sudoku.

\begin{figure}[t]
\vspace{-0.1cm}
\centering
\includegraphics[width=\textwidth]{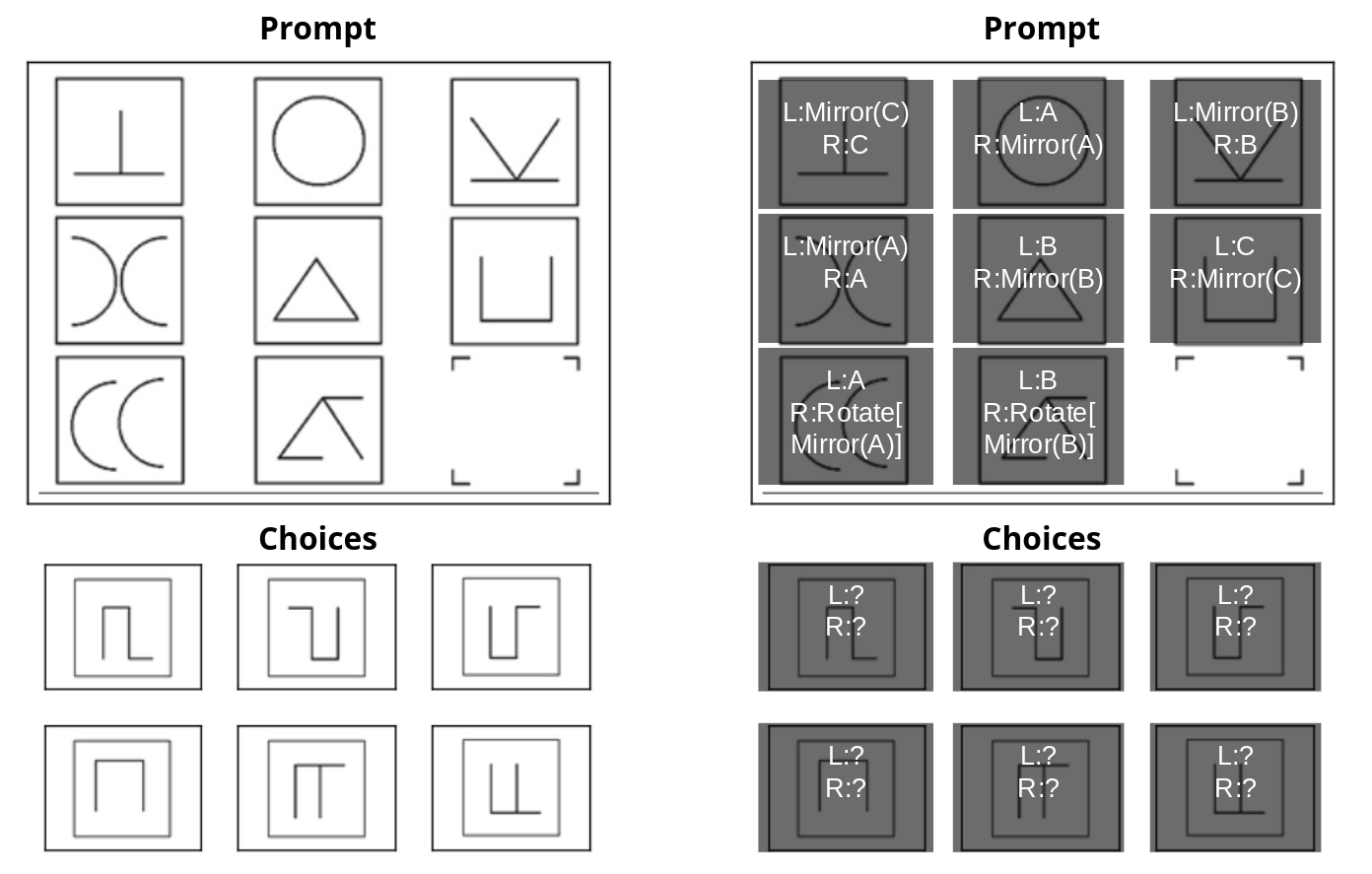}
\vspace{-0.5cm}
\caption{A challenging Raven's Matrix puzzle that exemplifies a difficult instance of the symbol grounding problem. We invite the reader to attempt the puzzle for themselves on the left hand side of the figure first, before looking at the annotations on the right hand side. Once the given images have been decoded to an appropriate symbolic representation, it is straightforward for a discrete solver or a human to solve it. For a full explanation of the solution, please see Appendix Section \ref{section:appendix_solution}.}
\label{fig:rpm_1}
\vspace{-0.1cm}
\end{figure}

The general complexity of the symbol grounding problem embedded in end-to-end learning should not be underestimated. Figure \ref{fig:rpm_1} directly exemplifies the difficulty of the symbol grounding problem for both human and artificial intelligence. Common measures of abstract reasoning in artificial intelligence such as DeepMind's PGM work similarly to Raven's Progressive Matrices (a test for human intelligence), where predicting what comes next involves determining the hidden attributes (symbols) in what has been presented (perceptual phenomena), and inferring the pattern from them \citep{chollet2019measure,zhang2019raven,barrett2018measuring,hu2020hierarchical}. Once given the hidden attributes, it is trivial for a human or a combinatorial solver to infer the pattern \citep{zhang2019raven}. However, jointly inferring the hidden attributes together with the pattern proves to be a challenging cognitive task in general.

% [move this]
% This might not be surprising, because it is common for negative results to not be reported. However, the deep learning community has increasingly found utility in so doing whenever the modes of failure can be properly identified, as shown recently in the case of adversarial defense \citep{athalye2018obfuscated,uesato2018adversarial}, the Adam optimizer \citep{wilson2017marginal,reddi2019convergence}, and meta reinforcement learning \citep{liu2019taming,foerster2018dice}.

\subsection{Our Contribution}
%We answer the above question in the negative: SATNet solves visual Sudoku using end-to-end gradient-based optimization, but not in an end-to-end learning fashion. 
% Despite SATNet's impressive achievements, to our knowledge, after more than a year later, there has been no direct follow-up work. Even though the authors open-sourced their code and SAT (being NP-complete) covers a huge class of practical problems, there is no subsequent application of SATNet to be found in the literature. We argue that this is primarily because SATNet, in its current form, does not meet the strong notion of end-to-end learning.
In this paper, our principal contribution is a re-assessment of SATNet that clarifies the extent of its capabilities and a discussion of practical solutions that will help future researchers train SATNet layers in deep networks.

First, we observed from the SATNet authors' open-source code that intermediate labels are leaked in the SATNet training process for visual Sudoku. The leaked labels essentially result in a two-step training process for SATNet, where it first uses the leaked labels to train a digit classifier, and then uses the symbolic representations of the digits to solve for the Sudoku puzzle. After removing the intermediate labels, SATNet was observed to completely fail at visual Sudoku (0\% test accuracy). If intermediate labels are available, it is possible to separately pre-train a digit classifier and then use SATNet, independent of a deep network, to solve for the puzzle. This might even be preferable, given our finding that SATNet fails in 8 out of 10 random seeds despite access to the labels, which is evidence that SATNet struggles to learn to ground the Sudoku digits into their symbolic representation. To be clear, the label leakage did not affect SATNet in the non-visual case, and its success on purely symbolic inputs and outputs nonetheless marks progress in ILP, but does not fix the field's persisting deficiency in dealing with perceptual input.

While solving \emph{difficult} instances of the symbol grounding problem like visual Sudoku or PGM might be beyond the reach of SATNet, we found that SATNet also cannot solve \emph{easy} instances, unless properly configured. We devised a test called the \emph{MNIST mapping problem}, whose solution requires merely digit classification (a simple problem for neural networks) and learning a bijective mapping between logical variables (a simple problem for discrete solvers). This test serves as an easy instance of the symbol grounding problem, and is suitable as a sanity test not just for SATNet, but other prospective differentiable symbolic solvers. Even on a simple test like this, a naive application of SATNet can cause it to perform worse than models without logical reasoning capabilities.

Our work identifies several factors that affect the learning dynamics of SATNet and provides practical suggestions for configuring SATNet to enable successful training. We reveal surprising complexities that are unique to SATNet and break standard deep learning norms. For example, using different learning rates for different layers in neural networks is not a common practice, since the use of Adam usually suffices. But for the case of SATNet, even when Adam is used, the backbone layer has to learn at a slower rate than the SATNet layer for successful training to occur. Surprisingly, we found that unconditionally increasing the number of auxiliary variables does not increase the
expressivity of the model, but instead leads to a complete failure in learning.  Further adjusting the choice of optimizer and neural architecture led to statistically significant improvements, culminating in near perfect test accuracy (99\%).

The rest of the paper is organized as follows: Section \ref{section:technical_background} reviews the relevant technical background for SATNet and visual Sudoku. Section \ref{section:bug_explanation} examines the subtle nature of the label leakage in the original SATNet paper and its ramifications. Section \ref{section:mapping_problem} describes the MNIST mapping problem, and investigates optimal SATNet configurations for this simple MNIST-based test. Finally, we conclude in Section \ref{section:discussion_conclusion}.

\section{Background}
\label{section:technical_background}
\subsection{SATNet}
SATNet is a neural network layer that solves a semidefinite programming (SDP) relaxation of the following MAXSAT problem,
\begin{equation}
\begin{split}
\max_{\tilde{v} \in\{-1,1\}^{n}} \sum_{j=1}^{m} \bigvee_{i=1}^{n} \mathbf{1}\left\{\tilde{s}_{i j} \tilde{v}_{i}>0\right\},
\end{split}
\end{equation}
where $\tilde{v} \in\{-1,1\}^{n}$ denotes assignments to $n$ binary variables, and $\tilde{s}_{i} \in\{-1,0,1\}^{m}$ denotes the sign of variable $\tilde{v}_i$ in $m$ clauses. The set of $\tilde{s}_{i j}$, denoted by $S$, forms the SATNet layer's learnable parameters. $\tilde{v}$ can be partitioned into two disjoint sets $\mathcal{I}$ and $\mathcal{O}$, which are represented in SATNet by layer inputs $Z_\mathcal{I}$ and outputs $Z_\mathcal{O}$ (which can be either probabilistic or strictly binary), and their respective continuous relaxations $V_\mathcal{I}$ and $V_\mathcal{O}$. Gradients from the layer output $\nabla_{Z_\mathcal{O}} \mathcal{L}$ are backpropagated to both the layer's weights in the form of $\nabla_S \mathcal{L}$ and to the layer input in the form of $\nabla_{Z_\mathcal{I}} \mathcal{L}$. The two main tunable hyperparameters in a SATNet layer are the number of clauses $m$ and the number of auxiliary variables $aux$ (which ``play a role akin to register memory that is useful for inference''). Auxiliary variables are also input variables, but unlike $Z_\mathcal{I}$, they are not the output of preceding layers.
\subsection{Visual Sudoku}
Sudoku is a number puzzle played out on a 9-by-9 grid. Each of the 9x9=81 cells has to contain a digit from 1 to 9. The game starts out from a partially filled grid, and the object of the game is to complete the rest of the cells on the grid. Each of the digits from 1 to 9 has to appear exactly once in every row, column, and each of the nine 3-by-3 subgrids. In the \emph{non-visual} case, the state of the Sudoku grid can be encoded using 9x81=729 binary variables, and SATNet can learn to map from the binary encoding of the initial grid to the binary encoding of the completed grid without the programmer having to explicitly encode for the rules of the game. Given 9000 training and 1000 test examples (with 36.2 pre-filled cells on average), where each example is a pair consisting of the initial and completed grid, SATNet achieves 99.7\% training and 98.3\% test accuracy. By comparison, a symbolic solver that knows the rules of the game can provably solve the game perfectly \citep{sudoku2006norvig}, while a purely deep learning based approach, trained on a million examples, scores 70.0\% on a test set of thirty games \citep{sudoku2018park}. We report on other related work on non-visual Sudoku in Appendix Section \ref{section:appendix_nonvisual}.

In \emph{visual} Sudoku, the inputs are now 81 images of digits (taken from the MNIST dataset), with `0' standing in for empty cells. They are processed by a convolutional neural network (CNN) backbone with a SATNet layer, which performs at 93.6\% training and 63.2\% test accuracy using the same number of training and test examples. The SATNet authors contextualized their findings by claiming that the ``theoretical best'' test accuracy is capped at 74.8\% ($\approx$ 0.992\textsuperscript{36.2}), which is the probability that the LeNet\footnote{To be precise, the SATNet authors used a bigger version with $\sim$10x more parameters than the original.}  CNN backbone, which has 99.2\% test accuracy on MNIST, has correctly classified all the pre-filled cells.

% [remove/reword this]
% \subsection{Symbol Grounding}
% The canonical thought experiment for the symbol grounding problem is the Chinese Dictionary argument, a close cousin of the more well-known Chinese Room argument \citep{searle1980minds}. Suppose you were trying to learn Chinese as a second language, but had access to only a Chinese dictionary. Then the self-contained nature of the hyperlinks within the dictionary would make it seem like every symbol is connected to another symbol in some meaningless fashion. Ultimately, meaning can only be achieved by grounding the symbols in a first language and real-world experience. Suppose now that Chinese was to be learned as a first language. Then, learning becomes impossible unless there is some kind of mechanism that directly grounds the internal cognition of the symbols to something external.

\section{SATNet Fails at Symbol Grounding}
\label{section:bug_explanation}
\subsection{The Absence of Output Masking}
%The absence of output masking is responsible for leaking intermediate labels in the SATNet author's Sudoku solution.

While every Sudoku puzzle corresponds to 729 logical variables in the MAXSAT problem (excluding the auxiliary variables for now), the number of pre-filled cells and their positions differ depending on the puzzle. Thus, $\mathcal{I}$ and $\mathcal{O}$ are different for each example, even though the sizes of $Z_\mathcal{I}$ and $Z_\mathcal{O}$ are fixed beforehand and not example-dependent. A straightforward way to solve this is to apply an appropriate bit mask depending on the example.

Consider a toy example with 5 variables $v_1 = 1, v_2 = 0, v_3 = 0, v_4 = 1, v_5 = 0$ where $\mathcal{I} = \{1,2,3\}$ and $\mathcal{O} = \{4,5\}$. Then, the input to SATNet should be $10000$ with the bit mask $11100$, and the output should be $00010$ with the bit mask $00011$. The problem with the original SATNet implementation is that the bits that correspond to the inputs are not masked in the output.

Not masking the output might not seem problematic, given that SATNet does not modify input variables $Z_\mathcal{I}$ nor their relaxations $V_\mathcal{I}$. But consider the decomposition of the loss function $\mathcal{L}$ into a sum of binary cross entropies (BCE) between the SATNet variables $z$ and the training label $l$.
\begin{equation}
\label{eqn:bce}
\begin{split}
\mathcal{L} = \sum^n_{i=1} \text{BCE}(z_i, l_i) = \sum_{i \in \mathcal{I}} \text{BCE}(z_i, l_i) + \sum_{o \in \mathcal{O}} \text{BCE}(z_o, l_o).
\end{split}
\end{equation}
Since the $z_\mathcal{I}$ are not modified by SATNet, $z_i = l_i$ for $i \in \mathcal{I}$, effectively zero-ing out any loss contributed by terms in $z_\mathcal{I}$. This is true when SATNet is applied to purely symbolic problems like non-visual Sudoku. 

\begin{figure}[t]
\vspace{-0.1cm}
\centering
\includegraphics[width=\textwidth]{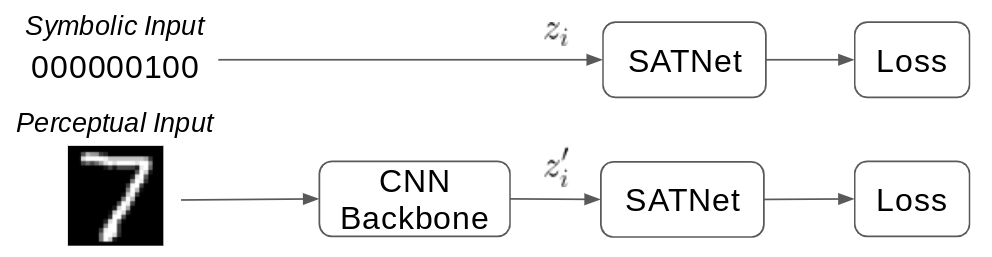}
\vspace{-0.5cm}
\caption{A visualization of the difference between symbolic and perceptual inputs.}
\label{fig:symbol_grounding_2}
% \vspace{-0.3cm}
\end{figure}

However, once perceptual input is introduced, $z_i$ is not directly accessible by SATNet. Instead, the input to the SATNet layer is a symbolic representation $z_i'$ of features extracted from the data (see Figure \ref{fig:symbol_grounding_2}). Thus, the loss from $z_\mathcal{I}$ in Equation \ref{eqn:bce} is non-zero before the neural network has learned to ground the symbols appropriately, i.e.~$z_i' = z_i = l_i$. Not masking the output to SATNet thus leaks label information to the layers before the SATNet layer, effectively training a classifier that learns to map from the perceptual data to the appropriate symbol representation, i.e.~symbol grounding.

\subsection{Visual Sudoku}
\begin{table}[h]
  \caption{Effects of Output Masking}
  \label{table:output_masking}
  \centering
  \begin{tabular}{lcccc}
    \toprule
    & \multicolumn{2}{c}{Non-Visual Sudoku} & \multicolumn{2}{c}{Visual Sudoku}                   \\
    \cmidrule(r){2-3} \cmidrule(r){4-5}
    Accuracy & Original & Masked Outputs & Original & Masked Outputs \\
    \midrule
    Train & 99.7$\pm$0.0\% & 99.7$\pm$0.0\% & 18.5$\pm$12.3\% & 0.0$\pm$0.0\% \\
    Test & 97.6$\pm$0.1\% & 97.6$\pm$0.1\% & 11.9$\pm$7.9\% & 0.0$\pm$0.0\% \\
    \bottomrule
  \end{tabular}
\end{table}
% visual:True, output_masked:True, train:train, mean:0.0, std:0.0
% visual:True, output_masked:True, train:test, mean:0.0, std:0.0
% visual:True, output_masked:False, train:train, mean:18.5, std:12.3
% visual:True, output_masked:False, train:test, mean:11.9, std:7.9
% visual:False, output_masked:True, train:train, mean:99.7, std:0.0
% visual:False, output_masked:True, train:test, mean:97.6, std:0.1
% visual:False, output_masked:False, train:train, mean:99.7, std:0.0
% visual:False, output_masked:False, train:test, mean:97.6, std:0.1
We re-ran the Sudoku experiments using the SATNet authors' open-sourced implementation with identical experimental settings, but over 10 different random seeds to get standard error confidence intervals. Table \ref{table:output_masking} shows clearly that output masking does not affect the results in the non-visual case, but causes SATNet to fail completely for visual Sudoku, which is what we expect from the discussion in the previous section. Once the intermediate labels are gone, the CNN does not ever learn to classify the digits better than chance. SATNet's failure at symbol grounding directly leads to its failure at the overall visual Sudoku task.

Interestingly, we also found that SATNet's performance in visual Sudoku in the absence of output masking is highly dependent on the random initialization, with 8/10 random seeds leading to complete failure as well. This explains why SATNet's performance over 10 runs (18.5\% training accuracy) is dramatically lower than what was originally reported (93.6\% training accuracy). Therefore, even for problems where we have access to intermediate labels, leaking them indirectly via the absence of output masking is strictly less desirable than directly pre-training a neural network classifier with those labels. In Section \ref{section:configuration}, we note important strategies for mitigating complete failure.

\begin{figure}[h]
\vspace{-0.1cm}
\centering
\includegraphics[width=\textwidth]{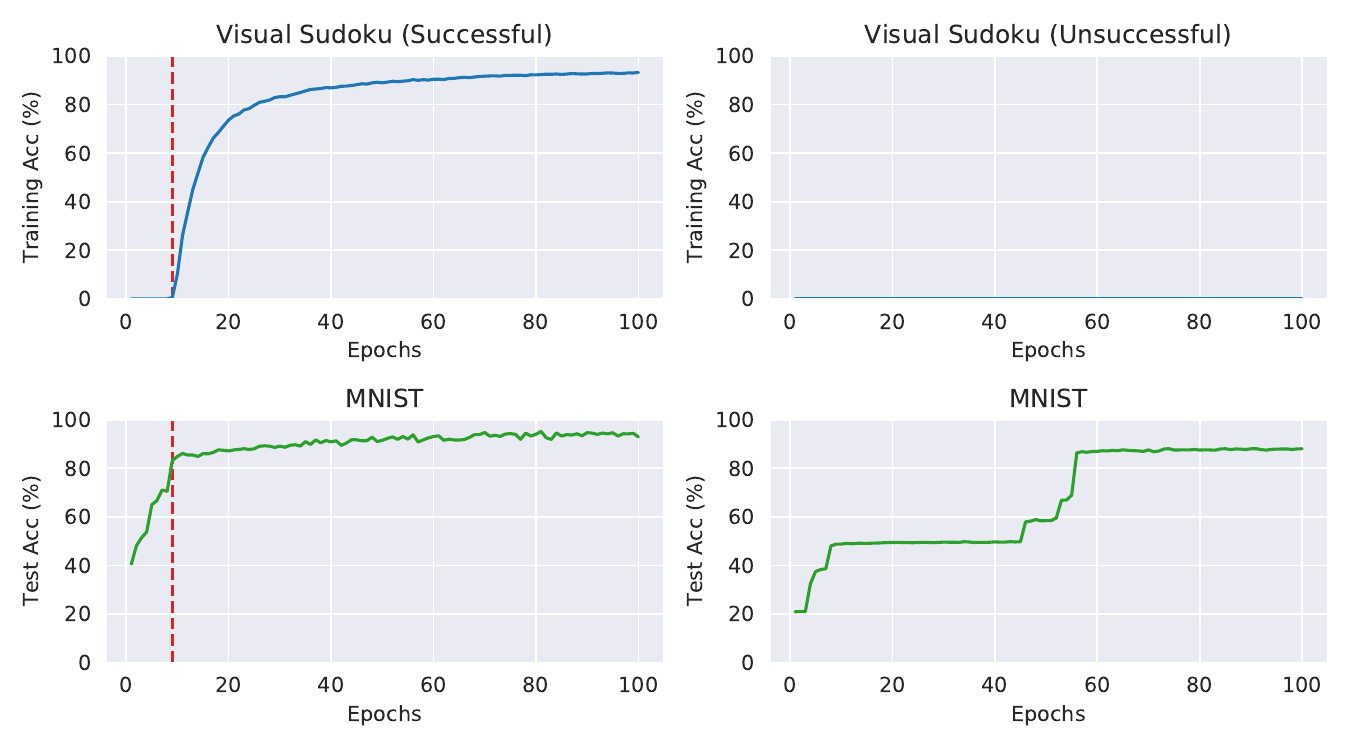}
\vspace{-0.5cm}
\caption{The graphs on the left show a successful run of SATNet on visual Sudoku, while the graphs on the right show an unsuccessful run. The successful run in the absence of output masking leads to a two-step training process, where the CNN first rapidly learns to classify digits, and then the SATNet layer learns to solve for Sudoku. The red vertical dotted line demarcates the point at which the training accuracy for visual Sudoku becomes non-zero. Unsuccessful runs typically take a long time for the CNN to classify digits, and never does better than 0\% training accuracy at the overall visual Sudoku task.}
\label{fig:symbol_grounding}
\vspace{-0.3cm}
\end{figure}

Of the 2 runs that succeeded (i.e.~had non-zero training accuracy, specifically 93.2\% and 91.7\% respectively), we found that the label leakage basically results in a two-step training process for SATNet, where the CNN first learns to do MNIST digit classification, and then the SATNet layer learns to solve the actual Sudoku problem. We show in Figure \ref{fig:symbol_grounding} training accuracy plots of two example runs, one successful and the other not. They are annotated with corresponding plots (at the bottom for comparison) of the CNN's classification accuracy on the MNIST test set. For the successful runs, we observe that the training accuracy for visual Sudoku stays at zero for a small number of epochs, during which time the leaked labels help train the CNN to be an MNIST digit classifier. Only after the digit classifier works to some degree, does the training accuracy for visual Sudoku actually become non-zero. By contrast, in most of the unsuccessful runs, the CNN takes a very long time to become somewhat proficient at digit classification, and even after it does so, the SATNet layer seems unable to adapt to it, resulting in a permanent plateau at 0\% training accuracy.

\section{MNIST Mapping Problem}
\label{section:mapping_problem}
The MNIST mapping problem involves a symbolic problem with 20 variables $v_i$, where the first ten variables are input (i.e.~$\mathcal{I}=\{1,\dots,10\}$), and the next ten are output (i.e.~$\mathcal{O}=\{11,\dots,20\}$). But the $v_{\mathcal{I}}$ are not provided directly; instead the input is given as perceptual data in the form of an MNIST digit image, and the challenge is to map an image of digit $i$ to the variable $v_{11+i}$. We assume that these variables are boolean (or the probabilistic equivalent, i.e.~random variables taking real values in $[0,1]$), but this should be adapted accordingly to the symbolic representation of a given solver.

There are two distinct sub-problems. The first sub-problem involves classifying an MNIST digit image into $v_1, \dots, v_{10}$ (using a neural network). The second sub-problem involves learning a bijection (or an equivalent permutation) to $v_{11}, \dots, v_{20}$ (using a symbolic solver), from which the class of the input image has to be identified. Both sub-problems taken on their own are considered to be \emph{easy} problems. MNIST digits can be easily classified to 99\% test accuracy \citep{lecun1998gradient}, while permutation groups under equivalence queries are known to be exactly learnable in polynomial time \citep{arvind1996complexity}. Hence, we propose that a suitable sanity test for a differentiable symbolic solver is to solve the MNIST mapping problem to an accuracy of 99\%. Note that a model that does not have to learn the bijection can circumvent the symbol grounding problem entirely by simply learning the output labels directly. Therefore, the test is strictly intended to be a check for symbol grounding, rather than a grand AI challenge that necessitates the combination of pattern recognition and logical reasoning as in visual Sudoku or PGM.

\subsection{Configuring SATNet Properly}
\label{section:configuration}
Surprisingly, some SATNet configurations fail the test, not by a slight margin, but completely (i.e.~test accuracy no better than chance; we count them using 12\% as a threshold to account for variance). In general, we found that the successful training of SATNet can be very sensitive to specific combinations of hyperparameters, optimizers, and neural architectures. We present four empirical findings using experiments on the MNIST mapping problem. All experiments were ran for 50 training epochs over 10 random seeds to get standard error confidence intervals. The Sudoku CNN, which was the backbone architecture used in the SATNet author's visual Sudoku implementation, is used throughout unless stated otherwise. We evaluate the results by presenting test accuracies with their confidence intervals and the number of complete failures in parentheses. For comparison, a non-SATNet baseline, which consists of the Sudoku CNN but with the SATNet layer replaced by two fully connected layers (1000 hidden units and ReLU), performs at 72.1$\pm$13.3\%~(3). At a minimum, SATNet should perform better than that, since its raison d'être disappears if it can be bested by equivalent models without logical reasoning capabilities.

\paragraph{Finding 1} \textit{Too little ``logic'' (i.e.~low $m$) or too much ``slack'' (i.e.~high $aux$) can cause failure.}

\begin{figure}[h]
\vspace{-0.1cm}
\centering
\includegraphics[width=\textwidth]{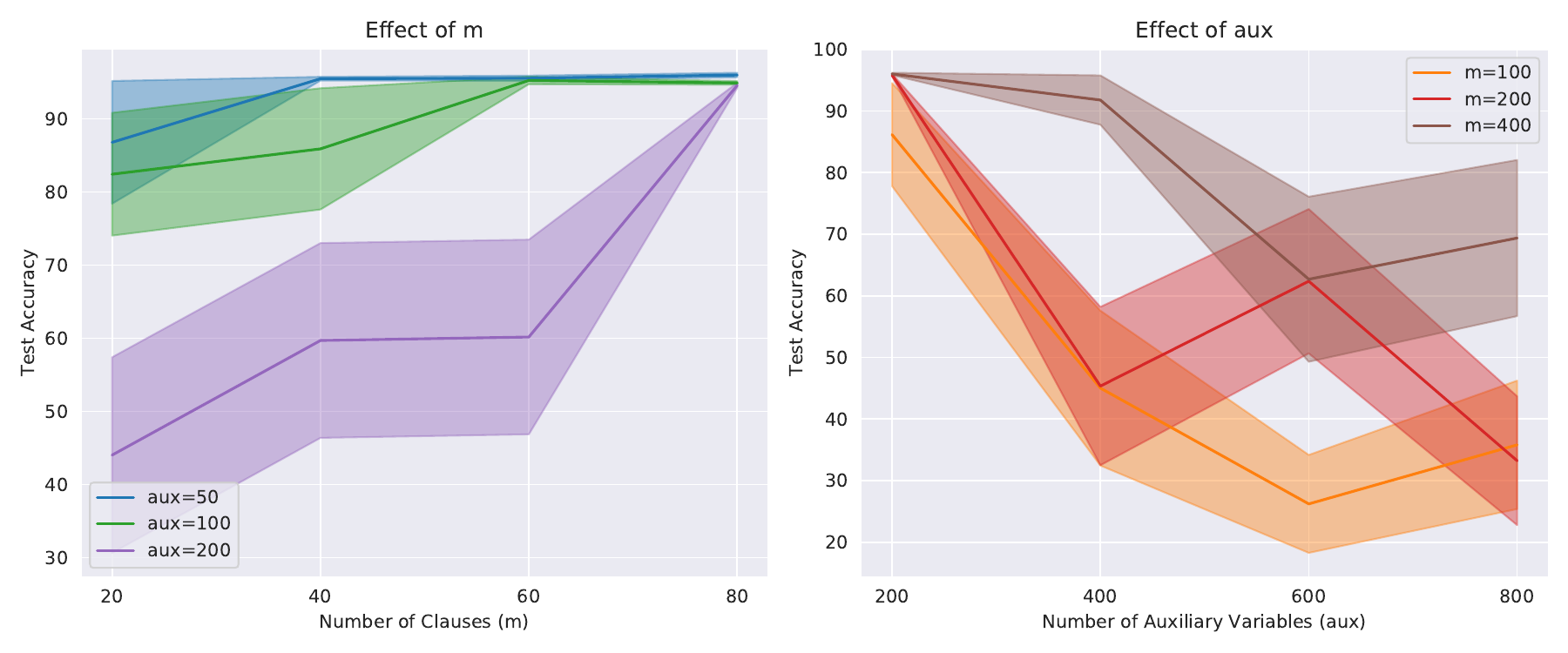}
\vspace{-0.5cm}
\caption{Both graphs show test accuracy on the MNIST mapping problem with the shaded interval representing the standard error.}
\label{fig:m_aux}
% \vspace{-0.3cm}
\end{figure}

The number of clauses $m$ controls the capacity of SATNet (rank of clause matrix), and we found that it can cause failure or result in terrible test accuracy when it is too low relative to what is needed for the problem. The number of auxiliary variables $aux$ also controls model capacity, but we observed that if it is too high for a given $m$, it can also cause failure (because most of the clauses end up being filled with meaningless input-independent auxiliary variables). High $m$ or low $aux$ do not affect test accuracy on the MNIST mapping problem, but they affect the amount of compute the SATNet layer uses.

\clearpage
\paragraph{Finding 2} \textit{The backbone layer has to learn at a slower rate than the SATNet layer.}
% SATNet lr:0.001000 Backbone lr:0.001000 Mean:19.9 Std:8.6 Complete Fails:9
% SATNet lr:0.001000 Backbone lr:0.000100 Mean:90.0 Std:8.7 Complete Fails:1
% SATNet lr:0.001000 Backbone lr:0.000010 Mean:96.3 Std:0.2 Complete Fails:0
% SATNet lr:0.000100 Backbone lr:0.001000 Mean:17.4 Std:4.3 Complete Fails:8
% SATNet lr:0.000100 Backbone lr:0.000100 Mean:74.6 Std:8.6 Complete Fails:0
% SATNet lr:0.000100 Backbone lr:0.000010 Mean:96.1 Std:0.2 Complete Fails:0
% SATNet lr:0.000010 Backbone lr:0.001000 Mean:14.8 Std:3.6 Complete Fails:9
% SATNet lr:0.000010 Backbone lr:0.000100 Mean:31.7 Std:7.1 Complete Fails:5
% SATNet lr:0.000010 Backbone lr:0.000010 Mean:72.4 Std:5.3 Complete Fails:0
\begin{table}[h]
  \caption{Effects of Different Learning Rates on the SATNet and Backbone Layer on Test Accuracy}
  \label{table:optimizer_lr_test}
  \centering
  \begin{tabular}{cccc}
    \toprule
    SATNet Layer & \multicolumn{3}{c}{Backbone Layer Learning Rate}\\
    \cmidrule(r){2-4}
    Learning Rate & 1x10\textsuperscript{-3} & 1x10\textsuperscript{-4} & 1x10\textsuperscript{-5} \\
    \midrule
    1x10\textsuperscript{-3} & 19.9$\pm$8.6\%~(9) & 90.0$\pm$8.7\%~(1) & 96.3$\pm$0.2\%~(0) \\
    1x10\textsuperscript{-4} & 17.4$\pm$4.3\%~(8) & 74.6$\pm$8.6\%~(0) & 96.1$\pm$0.2\%~(0) \\
    1x10\textsuperscript{-5} & 14.8$\pm$3.6\%~(9) & 31.7$\pm$7.1\%~(5) & 72.4$\pm$5.3\%~(0) \\
    \bottomrule
  \end{tabular}
\end{table}

Table \ref{table:optimizer_lr_test} shows the effect of differential learning rates between the SATNet and CNN backbone layers on test accuracy and number of failures, using Adam \citep{kingma2014adam} for both layers. If the backbone layer has a higher learning rate than the SATNet layer, this often leads to failure. Optimal performance is observed when the backbone layer has a lower learning rate than the SATNet layer. Note that this might be counter-intuitive, given that in the label leakage scenario, the backbone CNN had to learn digit recognition before the SATNet layer could learn to solve Sudoku. But without label leakage, having a higher learning rate for the backbone does not make sense because it cannot learn anything useful without the help of the SATNet layer.

\paragraph{Finding 3} \textit{Optimizing the backbone layer with SGD and the SATNet layer with Adam improves both training and test accuracy.}

Instead of simply using different learning rates, swapping the optimizer for the backbone layer with SGD raises test accuracy from 96.3 $\pm$0.2\%~(0) to 98.6$\pm$0.1\%~(0) (similarly so for training accuracy).

\paragraph{Finding 4} \textit{A sigmoid output layer for the backbone is preferable to softmax.}
\begin{table}[h]
  \caption{Effects of Different Neural Architectures on Test Accuracy}
  \label{table:architecture_test}
  \centering
  \begin{tabular}{ccccc}
    \toprule
    & & \multicolumn{2}{c}{Backbone Output Layer} \\
    \cmidrule(r){3-4}
    Architectures & Parameters & Softmax & Sigmoid \\
    \midrule
    LeNet \citep{lecun1998gradient} & 68,626 & 63.3$\pm$14.1\%~(4) & 98.8$\pm$0.0\%~(0)\\
    Sudoku CNN & 860,780 & 98.6$\pm$0.1\%~(0) & \textbf{99.1$\pm$0.0\%~(0)}\\ 
    ResNet18 \citep{he2016deep} & 11,723,722 & 67.6$\pm$6.3\%~(0) & 97.2$\pm$0.9\%~(0)\\
    \bottomrule
  \end{tabular}
\end{table}
% [63.25599999999999, 98.641, 67.64699999999999, 98.77699999999999, 99.08500000000001, 97.2]
% [14.127842423935636, 0.07651361534611634, 6.288719716020636, 0.04090775530928614, 0.016210421887717166, 0.8788123298583782]
% [4, 0, 0, 0, 0, 0]
% [68626, 860780, 11723722, 68626, 860780, 11723722]

The output of the CNN backbone has to take real values in $[0,1]$; the SATNet authors' implementation used a softmax output layer to achieve this. We found that a sigmoid output layer strictly outperforms a softmax layer in all three architectures tested. When softmax is used, we observed that the size of the architecture can result in poor performance if it is too small or too big. In the case where it is too big, it is possible for accuracy to degrade rapidly after reaching its peak (we don't use early stopping). Of the three, the Sudoku CNN appears to be the optimal size.

Every model we tested failed at visual Sudoku, demonstrating the non-triviality of visual Sudoku's grounding problem (since getting even one puzzle in the test set correct necessitates the accurate classification of 36.2 digits on average). However, even for a seemingly easy instance of the symbol grounding problem in the form of MNIST mapping, it was highly non-trivial to find the correct SATNet configuration that would lead to 99\% test accuracy. This shows that the current state of SATNet falls significantly short of its promise to integrate logical reasoning in deep learning.

\section{Conclusion}
\label{section:discussion_conclusion}
In this paper, we presented a detailed analysis of SATNet's capabilities, and provided practical solutions that will help future researchers train SATNet layers in their deep neural networks more effectively. Specifically, we noted that the original experimental setup for visual Sudoku resulted in intermediate label leakage. After removing the intermediate labels, SATNet is found to completely fail at the task of visual Sudoku due to its inability to ground the images of the puzzle digits into the appropriate symbolic representation. We further introduced the MNIST mapping problem as an easier instance of the symbol grounding problem compared to visual Sudoku, and found that SATNet needs to be delicately configured for training to be successful. In particular, the number of auxiliary variables cannot be increased unconditionally with respect to the number of clauses, and the backbone layer has to learn at a slower rate than the SATNet layer.

We can apply what we have learned about SATNet and its failure to solve visual Sudoku's symbol grounding problem more generally to other attempts to integrate logical reasoning into deep learning. Given that logical reasoning modules act at a symbolic level, while generic deep learning modules act at a sub-symbolic level, the interface between these two levels has to involve a symbol grounding problem. Hence, even if the intermediate label leakage identified in this paper might be SATNet-specific, we think that explicit tests against simple, interpretable instances of the symbol grounding problem will be fruitful for future researchers in discerning their claims about end-to-end learning (versus end-to-end gradient-based optimization). %like visual Sudoku or PGM.

In general, we think that the differences between deep learning and logic mirror the ones between continuous and discrete optimization. These differences go far deeper than the superficial lack of derivatives in discrete optimization, and we believe true progress has to come from significantly tighter integrations between deep learning and logic. We are excited that our work brings these differences to the forefront and encourages the community to think more critically about how to go about integrating logical reasoning into deep learning.

\section*{Broader Impact}
\paragraph{Reproducibility}
In recent years, there has been a reproducibility ``crisis'' in the natural sciences and medicine \citep{baker2016reproducibility,ioannidis2005most,fanelli2018opinion}, with the problem even extending into the computational sciences like machine learning \citep{mcdermott2019reproducibility,klein2018towards,colas2019hitchhiker,ghanta2018systems}. There is little incentive for independent researchers to put in the effort to re-verify the claims of a paper that has gone through peer review. This is not least because of the possibility that the failure in replication might be due to problems with the replication rather than problems with the original claims. However, we believe that prominent papers, especially ones like SATNet that have won conference awards, deserve extra scrutiny. By re-assessing SATNet's original claims, we provide additional credibility for established findings in the machine learning literature. Sober re-assessments of cutting edge AI technology also help to downplay the `hype', allowing non-expert stakeholders from the broader society to be clear-eyed about the current state of the art. We regret if this paper appears overly critical of the impressive achievements made by SATNet. A potentially negative consequence of our paper is that it discourages researchers from making their code open-source because of the additional scrutiny that this will invite. Critical assessments of AI technology might also lower both public and commercial funding for AI due to more realistic expectations, as has happened during the AI winters.

\paragraph{The Importance of the Symbol Grounding Problem}
There have been many attempts to combine pattern recognition and logical reasoning into a single neural network model, but most of these attempts essentially focus on reducing the problem to the relaxation of non-differentiable functions. Our work on SATNet clearly exemplifies that addressing the optimization issues inherent in combining logic and deep learning will not be enough to train models in a minimally supervised end-to-end learning fashion. Without a significant breakthrough, solving symbol grounding problems without intermediate labels will probably remain out of reach. Our work aims to highlight the importance of explicitly addressing the symbol grounding problem, and we hope that future research to do so will expand the applications of machine learning and AI beyond System-1 pattern recognition capabilities.

\section*{Funding Disclosure}
The work in this paper was done when the first author was an intern at Sony AI.

\clearpage

\bibliography{references}

\begin{thebibliography}{51}
\providecommand{\natexlab}[1]{#1}
\providecommand{\url}[1]{\texttt{#1}}
\expandafter\ifx\csname urlstyle\endcsname\relax
  \providecommand{\doi}[1]{doi: #1}\else
  \providecommand{\doi}{doi: \begingroup \urlstyle{rm}\Url}\fi

\bibitem[Wang et~al.(2019)Wang, Donti, Wilder, and Kolter]{wang2019satnet}
Po-Wei Wang, Priya~L Donti, Bryan Wilder, and Zico Kolter.
\newblock Satnet: Bridging deep learning and logical reasoning using a
  differentiable satisfiability solver.
\newblock \emph{arXiv preprint arXiv:1905.12149}, 2019.

\bibitem[Harnad(1990)]{harnad1990symbol}
Stevan Harnad.
\newblock The symbol grounding problem.
\newblock \emph{Physica D: Nonlinear Phenomena}, 42\penalty0 (1-3):\penalty0
  335--346, 1990.

\bibitem[Russakovsky et~al.(2015)Russakovsky, Deng, Su, Krause, Satheesh, Ma,
  Huang, Karpathy, Khosla, Bernstein, et~al.]{russakovsky2015imagenet}
Olga Russakovsky, Jia Deng, Hao Su, Jonathan Krause, Sanjeev Satheesh, Sean Ma,
  Zhiheng Huang, Andrej Karpathy, Aditya Khosla, Michael Bernstein, et~al.
\newblock Imagenet large scale visual recognition challenge.
\newblock \emph{International journal of computer vision}, 115\penalty0
  (3):\penalty0 211--252, 2015.

\bibitem[Xiong et~al.(2017)Xiong, Wu, Alleva, Droppo, Huang, and
  Stolcke]{xiong2017the}
Wayne Xiong, Lingfeng Wu, Fil Alleva, Jasha Droppo, Xuedong Huang, and Andreas
  Stolcke.
\newblock The microsoft 2017 conversational speech recognition system
  [technical report].
\newblock Technical Report MSR-TR-2017-39, Microsoft, August 2017.
\newblock URL
  \url{https://www.microsoft.com/en-us/research/publication/microsoft-2017-conversational-speech-recognition-system/}.

\bibitem[Amodei et~al.(2016)Amodei, Ananthanarayanan, Anubhai, Bai, Battenberg,
  Case, Casper, Catanzaro, Cheng, Chen, et~al.]{amodei2016deep}
Dario Amodei, Sundaram Ananthanarayanan, Rishita Anubhai, Jingliang Bai, Eric
  Battenberg, Carl Case, Jared Casper, Bryan Catanzaro, Qiang Cheng, Guoliang
  Chen, et~al.
\newblock Deep speech 2: End-to-end speech recognition in english and mandarin.
\newblock In \emph{International conference on machine learning}, pages
  173--182, 2016.

\bibitem[Silver et~al.(2016)Silver, Huang, Maddison, Guez, Sifre, Van
  Den~Driessche, Schrittwieser, Antonoglou, Panneershelvam, Lanctot,
  et~al.]{silver2016mastering}
David Silver, Aja Huang, Chris~J Maddison, Arthur Guez, Laurent Sifre, George
  Van Den~Driessche, Julian Schrittwieser, Ioannis Antonoglou, Veda
  Panneershelvam, Marc Lanctot, et~al.
\newblock Mastering the game of go with deep neural networks and tree search.
\newblock \emph{nature}, 529\penalty0 (7587):\penalty0 484, 2016.

\bibitem[Silver et~al.(2017)Silver, Hubert, Schrittwieser, Antonoglou, Lai,
  Guez, Lanctot, Sifre, Kumaran, Graepel, et~al.]{silver2017mastering}
David Silver, Thomas Hubert, Julian Schrittwieser, Ioannis Antonoglou, Matthew
  Lai, Arthur Guez, Marc Lanctot, Laurent Sifre, Dharshan Kumaran, Thore
  Graepel, et~al.
\newblock Mastering chess and shogi by self-play with a general reinforcement
  learning algorithm.
\newblock \emph{arXiv preprint arXiv:1712.01815}, 2017.

\bibitem[Badia et~al.(2020)Badia, Piot, Kapturowski, Sprechmann, Vitvitskyi,
  Guo, and Blundell]{badia2020agent57}
Adri{\`a}~Puigdom{\`e}nech Badia, Bilal Piot, Steven Kapturowski, Pablo
  Sprechmann, Alex Vitvitskyi, Daniel Guo, and Charles Blundell.
\newblock Agent57: Outperforming the atari human benchmark.
\newblock \emph{arXiv preprint arXiv:2003.13350}, 2020.

\bibitem[Ecoffet et~al.(2020)Ecoffet, Huizinga, Lehman, Stanley, and
  Clune]{ecoffet2020first}
Adrien Ecoffet, Joost Huizinga, Joel Lehman, Kenneth~O Stanley, and Jeff Clune.
\newblock First return then explore.
\newblock \emph{arXiv preprint arXiv:2004.12919}, 2020.

\bibitem[Marcus(2020)]{marcus2020next}
Gary Marcus.
\newblock The next decade in ai: four steps towards robust artificial
  intelligence.
\newblock \emph{arXiv preprint arXiv:2002.06177}, 2020.

\bibitem[Chollet(2019)]{chollet2019measure}
Fran{\c{c}}ois Chollet.
\newblock The measure of intelligence.
\newblock \emph{arXiv preprint arXiv:1911.01547}, 2019.

\bibitem[Evans(1984)]{evans1984heuristic}
Jonathan St~BT Evans.
\newblock Heuristic and analytic processes in reasoning.
\newblock \emph{British Journal of Psychology}, 75\penalty0 (4):\penalty0
  451--468, 1984.

\bibitem[Kahneman(2011)]{kahneman2011thinking}
Daniel Kahneman.
\newblock \emph{Thinking, fast and slow}.
\newblock Macmillan, 2011.

\bibitem[Bengio(2019)]{bengio2019from}
Yoshua Bengio.
\newblock From system 1 deep learning to system 2 deep learning, 2019.
\newblock URL
  \url{https://slideslive.com/38922304/from-system-1-deep-learning-to-system-2-deep-learning}.
\newblock Conference on Neural Information Processing Systems.

\bibitem[Akkaya et~al.(2019)Akkaya, Andrychowicz, Chociej, Litwin, McGrew,
  Petron, Paino, Plappert, Powell, Ribas, et~al.]{akkaya2019solving}
Ilge Akkaya, Marcin Andrychowicz, Maciek Chociej, Mateusz Litwin, Bob McGrew,
  Arthur Petron, Alex Paino, Matthias Plappert, Glenn Powell, Raphael Ribas,
  et~al.
\newblock Solving rubik's cube with a robot hand.
\newblock \emph{arXiv preprint arXiv:1910.07113}, 2019.

\bibitem[Hu et~al.(2016)Hu, Ma, Liu, Hovy, and Xing]{hu2016harnessing}
Zhiting Hu, Xuezhe Ma, Zhengzhong Liu, Eduard Hovy, and Eric Xing.
\newblock Harnessing deep neural networks with logic rules.
\newblock \emph{arXiv preprint arXiv:1603.06318}, 2016.

\bibitem[Rockt{\"a}schel and Riedel(2017)]{rocktaschel2017end}
Tim Rockt{\"a}schel and Sebastian Riedel.
\newblock End-to-end differentiable proving.
\newblock In \emph{Advances in Neural Information Processing Systems}, pages
  3788--3800, 2017.

\bibitem[Cingillioglu and Russo(2018)]{cingillioglu2018deeplogic}
Nuri Cingillioglu and Alessandra Russo.
\newblock Deeplogic: Towards end-to-end differentiable logical reasoning.
\newblock \emph{arXiv preprint arXiv:1805.07433}, 2018.

\bibitem[Evans and Grefenstette(2018)]{evans2018learning}
Richard Evans and Edward Grefenstette.
\newblock Learning explanatory rules from noisy data.
\newblock \emph{Journal of Artificial Intelligence Research}, 61:\penalty0
  1--64, 2018.

\bibitem[Serafini and Garcez(2016)]{serafini2016logic}
Luciano Serafini and Artur~d'Avila Garcez.
\newblock Logic tensor networks: Deep learning and logical reasoning from data
  and knowledge.
\newblock \emph{arXiv preprint arXiv:1606.04422}, 2016.

\bibitem[Sourek et~al.(2015)Sourek, Aschenbrenner, Zelezny, and
  Kuzelka]{sourek2015lifted}
Gustav Sourek, Vojtech Aschenbrenner, Filip Zelezny, and Ondrej Kuzelka.
\newblock Lifted relational neural networks.
\newblock \emph{arXiv preprint arXiv:1508.05128}, 2015.

\bibitem[van Krieken et~al.(2020)van Krieken, Acar, and van
  Harmelen]{van2020analyzing}
Emile van Krieken, Erman Acar, and Frank van Harmelen.
\newblock Analyzing differentiable fuzzy logic operators.
\newblock \emph{arXiv preprint arXiv:2002.06100}, 2020.

\bibitem[Vlastelica et~al.(2019)Vlastelica, Paulus, Musil, Martius, and
  Rol{\'\i}nek]{vlastelica2019differentiation}
Marin Vlastelica, Anselm Paulus, V{\'\i}t Musil, Georg Martius, and Michal
  Rol{\'\i}nek.
\newblock Differentiation of blackbox combinatorial solvers.
\newblock \emph{arXiv preprint arXiv:1912.02175}, 2019.

\bibitem[Rol{\'\i}nek et~al.(2020)Rol{\'\i}nek, Swoboda, Zietlow, Paulus,
  Musil, and Martius]{rolinek2020deep}
Michal Rol{\'\i}nek, Paul Swoboda, Dominik Zietlow, Anselm Paulus, V{\'\i}t
  Musil, and Georg Martius.
\newblock Deep graph matching via blackbox differentiation of combinatorial
  solvers.
\newblock \emph{arXiv preprint arXiv:2003.11657}, 2020.

\bibitem[Tschiatschek et~al.(2018)Tschiatschek, Sahin, and
  Krause]{tschiatschek2018differentiable}
Sebastian Tschiatschek, Aytunc Sahin, and Andreas Krause.
\newblock Differentiable submodular maximization.
\newblock \emph{arXiv preprint arXiv:1803.01785}, 2018.

\bibitem[Blondel et~al.(2020)Blondel, Teboul, Berthet, and
  Djolonga]{blondel2020fast}
Mathieu Blondel, Olivier Teboul, Quentin Berthet, and Josip Djolonga.
\newblock Fast differentiable sorting and ranking.
\newblock \emph{arXiv preprint arXiv:2002.08871}, 2020.

\bibitem[Amos and Kolter(2017)]{amos2017optnet}
Brandon Amos and J~Zico Kolter.
\newblock Optnet: Differentiable optimization as a layer in neural networks.
\newblock In \emph{Proceedings of the 34th International Conference on Machine
  Learning-Volume 70}, pages 136--145. JMLR. org, 2017.

\bibitem[Santoro et~al.(2017)Santoro, Raposo, Barrett, Malinowski, Pascanu,
  Battaglia, and Lillicrap]{santoro2017simple}
Adam Santoro, David Raposo, David~G Barrett, Mateusz Malinowski, Razvan
  Pascanu, Peter Battaglia, and Timothy Lillicrap.
\newblock A simple neural network module for relational reasoning.
\newblock In \emph{Advances in neural information processing systems}, pages
  4967--4976, 2017.

\bibitem[Palm et~al.(2018)Palm, Paquet, and Winther]{palm2018recurrent}
Rasmus Palm, Ulrich Paquet, and Ole Winther.
\newblock Recurrent relational networks.
\newblock In \emph{Advances in Neural Information Processing Systems}, pages
  3368--3378, 2018.

\bibitem[Selsam et~al.(2018)Selsam, Lamm, B{\"u}nz, Liang, de~Moura, and
  Dill]{selsam2018learning}
Daniel Selsam, Matthew Lamm, Benedikt B{\"u}nz, Percy Liang, Leonardo de~Moura,
  and David~L Dill.
\newblock Learning a sat solver from single-bit supervision.
\newblock \emph{arXiv preprint arXiv:1802.03685}, 2018.

\bibitem[Manhaeve et~al.(2018)Manhaeve, Dumancic, Kimmig, Demeester, and
  De~Raedt]{manhaeve2018deepproblog}
Robin Manhaeve, Sebastijan Dumancic, Angelika Kimmig, Thomas Demeester, and Luc
  De~Raedt.
\newblock Deepproblog: Neural probabilistic logic programming.
\newblock In \emph{Advances in Neural Information Processing Systems}, pages
  3749--3759, 2018.

\bibitem[Yang et~al.(2017)Yang, Yang, and Cohen]{yang2017differentiable}
Fan Yang, Zhilin Yang, and William~W Cohen.
\newblock Differentiable learning of logical rules for knowledge base
  reasoning.
\newblock In \emph{Advances in Neural Information Processing Systems}, pages
  2319--2328, 2017.

\bibitem[Cropper et~al.(2020)Cropper, Duman{\v{c}}i{\'c}, and
  Muggleton]{cropper2020turning}
Andrew Cropper, Sebastijan Duman{\v{c}}i{\'c}, and Stephen~H Muggleton.
\newblock Turning 30: New ideas in inductive logic programming.
\newblock \emph{arXiv preprint arXiv:2002.11002}, 2020.

\bibitem[Cohen(2016)]{cohen2016tensorlog}
William~W Cohen.
\newblock Tensorlog: A differentiable deductive database.
\newblock \emph{arXiv preprint arXiv:1605.06523}, 2016.

\bibitem[Zhang et~al.(2019)Zhang, Gao, Jia, Zhu, and Zhu]{zhang2019raven}
Chi Zhang, Feng Gao, Baoxiong Jia, Yixin Zhu, and Song-Chun Zhu.
\newblock Raven: A dataset for relational and analogical visual reasoning.
\newblock In \emph{Proceedings of the IEEE Conference on Computer Vision and
  Pattern Recognition}, pages 5317--5327, 2019.

\bibitem[Barrett et~al.(2018)Barrett, Hill, Santoro, Morcos, and
  Lillicrap]{barrett2018measuring}
David~GT Barrett, Felix Hill, Adam Santoro, Ari~S Morcos, and Timothy
  Lillicrap.
\newblock Measuring abstract reasoning in neural networks.
\newblock \emph{arXiv preprint arXiv:1807.04225}, 2018.

\bibitem[Hu et~al.(2020)Hu, Ma, Liu, Wei, and Bai]{hu2020hierarchical}
Sheng Hu, Yuqing Ma, Xianglong Liu, Yanlu Wei, and Shihao Bai.
\newblock Hierarchical rule induction network for abstract visual reasoning.
\newblock \emph{arXiv preprint arXiv:2002.06838}, 2020.

\bibitem[Norvig(2006)]{sudoku2006norvig}
Peter Norvig.
\newblock Solving every sudoku puzzle.
\newblock \url{http://norvig.com/sudoku.html}, 2006.

\bibitem[Park(2018)]{sudoku2018park}
Kyubyong Park.
\newblock Can convolutional neural networks crack sudoku puzzles?
\newblock \url{https://github.com/Kyubyong/sudoku}, 2018.

\bibitem[LeCun et~al.(1998)LeCun, Bottou, Bengio, and
  Haffner]{lecun1998gradient}
Yann LeCun, L{\'e}on Bottou, Yoshua Bengio, and Patrick Haffner.
\newblock Gradient-based learning applied to document recognition.
\newblock \emph{Proceedings of the IEEE}, 86\penalty0 (11):\penalty0
  2278--2324, 1998.

\bibitem[Arvind and Vinodchandran(1996)]{arvind1996complexity}
Vikraman Arvind and NV~Vinodchandran.
\newblock The complexity of exactly learning algebraic concepts.
\newblock In \emph{International Workshop on Algorithmic Learning Theory},
  pages 100--112. Springer, 1996.

\bibitem[Kingma and Ba(2014)]{kingma2014adam}
Diederik~P Kingma and Jimmy Ba.
\newblock Adam: A method for stochastic optimization.
\newblock \emph{arXiv preprint arXiv:1412.6980}, 2014.

\bibitem[He et~al.(2016)He, Zhang, Ren, and Sun]{he2016deep}
Kaiming He, Xiangyu Zhang, Shaoqing Ren, and Jian Sun.
\newblock Deep residual learning for image recognition.
\newblock In \emph{Proceedings of the IEEE conference on computer vision and
  pattern recognition}, pages 770--778, 2016.

\bibitem[Baker(2016)]{baker2016reproducibility}
Monya Baker.
\newblock Reproducibility crisis?
\newblock \emph{Nature}, 533\penalty0 (26):\penalty0 353--66, 2016.

\bibitem[Ioannidis(2005)]{ioannidis2005most}
John~PA Ioannidis.
\newblock Why most published research findings are false.
\newblock \emph{PLos med}, 2\penalty0 (8):\penalty0 e124, 2005.

\bibitem[Fanelli(2018)]{fanelli2018opinion}
Daniele Fanelli.
\newblock Opinion: Is science really facing a reproducibility crisis, and do we
  need it to?
\newblock \emph{Proceedings of the National Academy of Sciences}, 115\penalty0
  (11):\penalty0 2628--2631, 2018.

\bibitem[McDermott et~al.(2019)McDermott, Wang, Marinsek, Ranganath, Ghassemi,
  and Foschini]{mcdermott2019reproducibility}
Matthew McDermott, Shirly Wang, Nikki Marinsek, Rajesh Ranganath, Marzyeh
  Ghassemi, and Luca Foschini.
\newblock Reproducibility in machine learning for health.
\newblock \emph{arXiv preprint arXiv:1907.01463}, 2019.

\bibitem[Klein et~al.(2018)Klein, Christiansen, Murphy, and
  Hutter]{klein2018towards}
Aaron Klein, Eric Christiansen, Kevin Murphy, and Frank Hutter.
\newblock Towards reproducible neural architecture and hyperparameter search.
\newblock \emph{Reproducibility in Machine Learning Workshop}, 2018.

\bibitem[Colas et~al.(2019)Colas, Sigaud, and Oudeyer]{colas2019hitchhiker}
C{\'e}dric Colas, Olivier Sigaud, and Pierre-Yves Oudeyer.
\newblock A hitchhiker's guide to statistical comparisons of reinforcement
  learning algorithms.
\newblock \emph{arXiv preprint arXiv:1904.06979}, 2019.

\bibitem[Ghanta et~al.(2018)Ghanta, Khermosh, Subramanian, Sridhar,
  Sundararaman, Arteaga, Luo, Roselli, Das, and Talagala]{ghanta2018systems}
Sindhu Ghanta, Lior Khermosh, Sriram Subramanian, Vinay Sridhar, Swaminathan
  Sundararaman, Dulcardo Arteaga, Qianmei Luo, Drew Roselli, Dhananjoy Das, and
  Nisha Talagala.
\newblock A systems perspective to reproducibility in production machine
  learning domain.
\newblock \emph{Reproducibility in Machine Learning Workshop}, 2018.

\bibitem[user265554
  (https://puzzling.stackexchange.com/users/16477/user265554)(2015)]{ravenpuzzle}
user265554 (https://puzzling.stackexchange.com/users/16477/user265554).
\newblock Complete the sequence.
\newblock Puzzling Stack Exchange, 2015.
\newblock URL
  \url{https://puzzling.stackexchange.com/questions/22495/complete-the-sequence}.
\newblock
  URL:https://puzzling.stackexchange.com/questions/22495/complete-the-sequence
  (version: 2020-05-08).

\end{thebibliography}
\bibliographystyle{unsrtnat}

\clearpage
\appendix
\section*{Appendix}
\section{Solution to the Raven's Matrix puzzle}
\label{section:appendix_solution}
\begin{figure}[ht]
\centering
\includegraphics[width=0.5\textwidth]{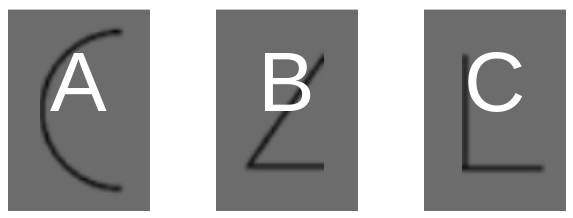}
\vspace{-0.1cm}
\caption{The three basic glyphs are formed from half a circle, a triangle, and a rectangle respectively.}
\label{fig:rpm_3}
\end{figure}
\begin{figure}[ht]
\centering
\includegraphics[width=\textwidth]{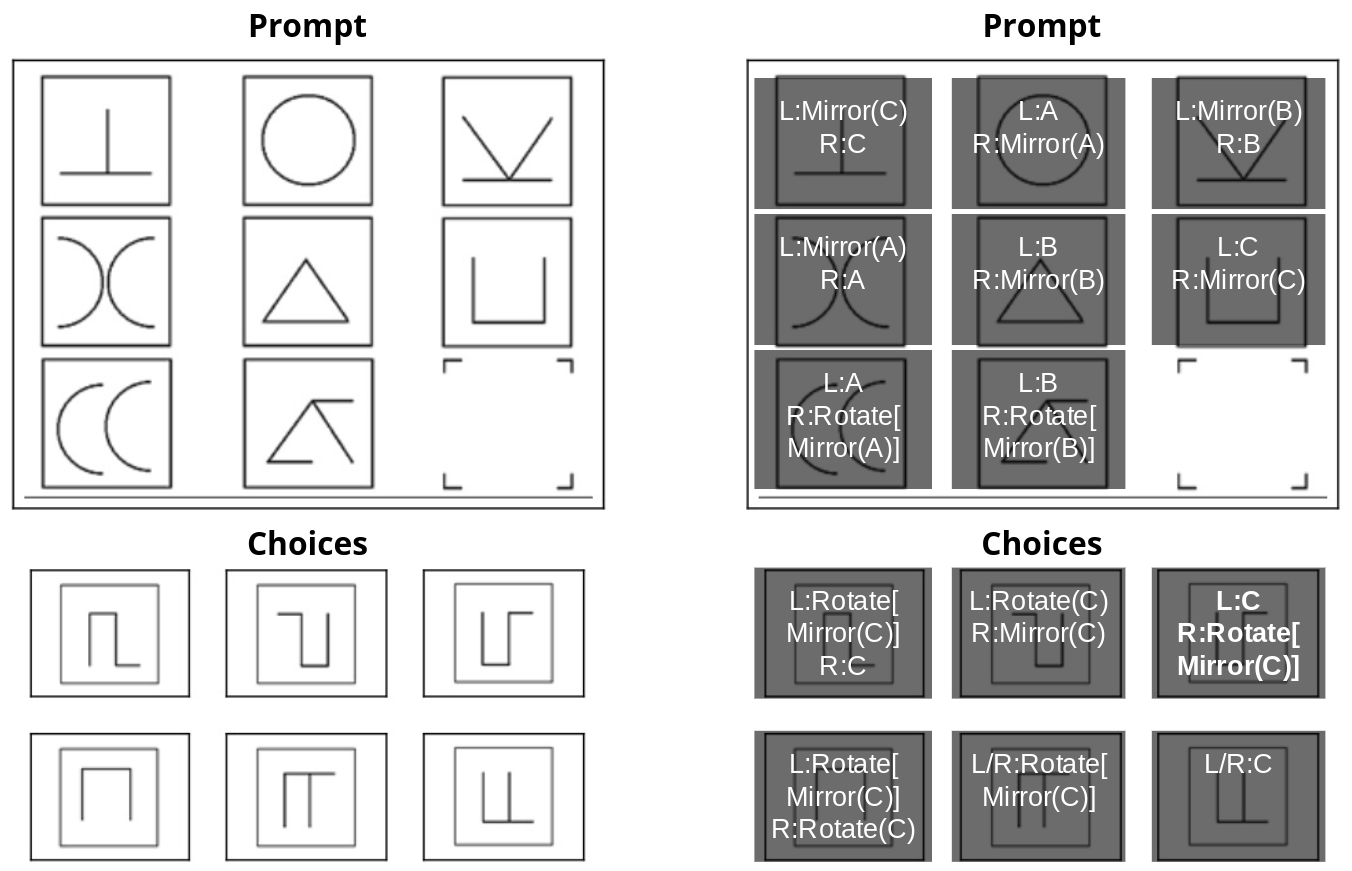}
\vspace{-0.4cm}
\caption{The solution to the Raven's Matrix puzzle is the choice on the top right.}
\label{fig:rpm_2}
\end{figure}

The source of this puzzle and its solution is \citet{ravenpuzzle} on Puzzling Stack Exchange.

Each panel is composed of a glyph on the left hand side (L) and a glyph on the right hand side (R). There are three basic glyphs (see Figure \ref{fig:rpm_3}): a crescent (A), a half triangle (B), and a half rectangle (C). Each glyph can also be mirrored (Mirror), i.e.~flipped horizontally, or rotated by 180 degrees (Rotate). In Figure \ref{fig:rpm_2}, we annotate every panel in both the prompt and the choices with the symbols that represent it. It is clear that the blank in the prompt should be filled by a left glyph C and a right glyph Rotate[Mirror(C)], which is the choice on the top right.

\section{Related Work on Non-Visual Sudoku}
\label{section:appendix_nonvisual}
On a dataset with 216,000 puzzles split in a 10:1:1 train-val-test ratio, a deep (recurrent relational) network that has access to positional information for each cell scores 100\% test accuracy on puzzles with 33 pre-filled cells and 96.6\% on puzzles with 17 pre-filled cells \citep{palm2018recurrent}. \citet{amos2017optnet} use a differentiable quadratic programming layer called OptNet, which like SATNet has no a priori knowledge of the rules, in a neural network to solve for Sudoku. OptNet does not scale well computationally and can only solve 4-by-4 Sudokus.

\section{Experimental Settings}
In the Supplementary materials, we provide source code and the shell commands to replicate all the experimental results in the paper.

\subsection{SATNet Fails at Symbol Grounding}
\label{section:appendix_symbol}
The experimental settings for SATNet in Section \ref{section:bug_explanation} are identical to the original paper and based on the authors' open-sourced implementation available at \url{https://github.com/locuslab/SATNet}. Specifically, the CNN used is the sequence of layers: \textit{Conv1-ReLU-MaxPool-Conv2-ReLU-MaxPool-FC1-ReLU-FC2-Softmax}, where \textit{Conv1} has a 5x5 kernel (stride 1) and 20 output channels, \textit{Conv2} has a 5x5 kernel (stride 1) and 50 output channels, \textit{FC1} has size 800x500, \textit{FC2} has size 500x10, and the \textit{MaxPool} layers have a 2x2 kernel (stride 2). This is roughly the LeNet5 architecture, but with one less fully connected layer at the end and around 10x the number of parameters. The SATNet layer contains 300 auxiliary variables, with $n=729$ and $m=600$. The full model is trained using Adam for 100 epochs using batch size 40, with a learning rate of 2x10\textsuperscript{-3} for the SATNet layer and 1x10\textsuperscript{-5} for the CNN. %Note that the model is permute-invariant so the permutation of the inputs do not affect the results. We run the experiments using 10 different random seeds to obtain standard error confidence intervals.

\subsection{MNIST Mapping Problem}
\label{section:appendix_mapping}
We use batch size 64 for training throughout all the experiments. We use the Sudoku CNN described above in Appendix Section \ref{section:appendix_symbol} as the backbone layer for all the experiments, except the one in Finding 4 where we vary the architecture. We use $m=200,aux=100$ for the SATNet layer for all the experiments, except the one in Finding 1 where we vary $m$ and $aux$.

Non-SATNet baseline: The whole network was trained with Adam using a 2x10\textsuperscript{-3} learning rate. 

Finding 1: The SATNet layer was trained with a 2x10\textsuperscript{-3} learning rate, and the backbone layer was trained with a 1x10\textsuperscript{-5} learning rate, both using Adam as was done above in Appendix Section \ref{section:appendix_symbol}.

Finding 2: Both the SATNet layer and the backbone layer were trained with Adam.

Findings 3 and 4: The SATNet layer was trained with a 1x10\textsuperscript{-3} learning rate using Adam, and the backbone layer was trained with a 1x10\textsuperscript{-1} learning rate with SGD.

\clearpage
\section{More Experimental Results for the MNIST Mapping Problem}
\label{section:appendix_moreresults}
\subsection{Non-SATNet Baseline}
The training accuracy for the non-SATNet baseline is 72.4$\pm$13.4\%~(3).

\subsection{Experiment 1}
\begin{table}[h]
  \caption{Effects of $m$ and $aux$ on Training and Test Accuracy}
  \label{table:m_aux_full_comparison}
  \centering
  \begin{tabular}{cccc}
    \toprule
    $m$ & $aux$ & Training Accuracy & Test Accuracy \\
    \midrule
    20 & 50 & 86.7$\pm$8.4\%~(1) & 86.8$\pm$8.4\%~(1) \\
    40 & 50 & 95.6$\pm$0.3\%~(0) & 95.5$\pm$0.3\%~(0) \\
    60 & 50 & 95.7$\pm$0.3\%~(0) & 95.6$\pm$0.4\%~(0) \\
    80 & 50 & 96.2$\pm$0.2\%~(0) & 96.0$\pm$0.3\%~(0) \\
    20 & 100 & 82.2$\pm$8.4\%~(1) & 82.4$\pm$8.4\%~(1) \\
    40 & 100 & 85.9$\pm$8.3\%~(1) & 85.9$\pm$8.3\%~(1) \\
    60 & 100 & 95.3$\pm$0.5\%~(0) & 95.3$\pm$0.5\%~(0) \\
    80 & 100 & 95.1$\pm$0.2\%~(0) & 94.9$\pm$0.2\%~(0) \\
    20 & 200 & 43.9$\pm$13.5\%~(6) & 44.0$\pm$13.4\%~(6) \\
    40 & 200 & 59.6$\pm$13.3\%~(4) & 59.7$\pm$13.3\%~(4) \\
    60 & 200 & 60.0$\pm$13.4\%~(4) & 60.2$\pm$13.3\%~(4) \\
    80 & 200 & 94.7$\pm$0.3\%~(0) & 94.6$\pm$0.3\%~(0) \\
    100 & 200 & 86.3$\pm$8.4\%~(1) & 86.2$\pm$8.4\%~(1) \\
    100 & 400 & 44.8$\pm$12.5\%~(4) & 45.0$\pm$12.6\%~(4) \\
    100 & 600 & 25.6$\pm$7.7\%~(7) & 26.2$\pm$7.9\%~(7) \\
    100 & 800 & 35.1$\pm$10.3\%~(6) & 35.8$\pm$10.4\%~(6) \\
    200 & 200 & 96.2$\pm$0.1\%~(0) & 95.8$\pm$0.2\%~(0) \\
    200 & 400 & 45.6$\pm$12.9\%~(4) & 45.3$\pm$12.9\%~(4) \\
    200 & 600 & 62.4$\pm$11.5\%~(2) & 62.4$\pm$11.7\%~(2) \\
    200 & 800 & 32.7$\pm$10.4\%~(5) & 33.2$\pm$10.5\%~(5) \\
    400 & 200 & 96.4$\pm$0.2\%~(0) & 96.0$\pm$0.2\%~(0) \\
    400 & 400 & 92.1$\pm$4.2\%~(0) & 91.8$\pm$4.0\%~(0) \\
    400 & 600 & 62.8$\pm$13.5\%~(3) & 62.7$\pm$13.4\%~(3) \\
    400 & 800 & 69.3$\pm$12.8\%~(3) & 69.4$\pm$12.7\%~(3) \\
    \bottomrule
  \end{tabular}
\end{table}

% Test
% Effect of m
% [20, 40, 60, 80]
% [[86.792 95.49  95.555 95.996]
%  [82.433 85.898 95.251 94.907]
%  [44.042 59.706 60.181 94.577]]
% [[ 8.38830175  0.25828064  0.35464928  0.30500346]
%  [ 8.39569149  8.28944333  0.51322391  0.24770525]
%  [13.37093837 13.3143611  13.30899065  0.34767817]]
% [1, 0, 0, 0, 1, 1, 0, 0, 6, 4, 4, 0]
% Effect of aux
% [200, 400, 600, 800]
% [[86.167 45.013 26.213 35.802]
%  [95.84  45.344 62.365 33.223]
%  [96.02  91.777 62.687 69.375]]
% [[ 8.36336735 12.57149386  7.94976199 10.4232378 ]
%  [ 0.19686431 12.8505847  11.71510831 10.45200524]
%  [ 0.18761663  4.00173519 13.39755973 12.67240723]]
% [1, 4, 7, 6, 0, 4, 2, 5, 0, 0, 3, 3]

% Train
% Effect of m
% [20, 40, 60, 80]
% [[86.7465     95.531      95.7495     96.15516667]
%  [82.23383333 85.91666667 95.2935     95.08116667]
%  [43.88016667 59.645      60.02116667 94.6715    ]]
% [[ 8.4051978   0.28427677  0.33838734  0.24805347]
%  [ 8.43451003  8.31373746  0.51149107  0.22695135]
%  [13.45978445 13.26865099 13.36502191  0.34108473]]
% [1, 0, 0, 0, 1, 1, 0, 0, 6, 4, 4, 0]
% Effect of aux
% [200, 400, 600, 800]
% [[86.27016667 44.80883333 25.557      35.12716667]
%  [96.24416667 45.59566667 62.37666667 32.732     ]
%  [96.43216667 92.06216667 62.813      69.2775    ]]
% [[ 8.37120283 12.52342528  7.67475876 10.33127985]
%  [ 0.14574103 12.93116609 11.54996526 10.41867595]
%  [ 0.16621012  4.17678605 13.46616973 12.75020673]]
% [1, 4, 7, 6, 0, 4, 2, 5, 0, 0, 3, 3]

\subsection{Experiment 2}
\begin{table}[h]
  \caption{Effects of Different Learning Rates on the SATNet and Backbone Layer on Training Accuracy}
  \label{table:optimizer_lr_train}
  \centering
  \begin{tabular}{cccc}
    \toprule
    SATNet Layer & \multicolumn{3}{c}{Backbone Layer Learning Rate}\\
    \cmidrule(r){2-4}
    Learning Rate & 1x10\textsuperscript{-3} & 1x10\textsuperscript{-4} & 1x10\textsuperscript{-5} \\
    \midrule
    1x10\textsuperscript{-3} & 19.6$\pm$8.5\%~(9) & 90.4$\pm$8.8\%~(1) & 96.7$\pm$0.2\%~(0) \\
    1x10\textsuperscript{-4} & 17.0$\pm$4.1\%~(8) & 74.9$\pm$8.8\%~(0) & 96.5$\pm$0.2\%~(0) \\
    1x10\textsuperscript{-5} & 14.4$\pm$3.4\%~(9) & 31.8$\pm$7.1\%~(5) & 71.9$\pm$5.4\%~(0) \\
    \bottomrule
  \end{tabular}
\end{table}
% SATNet lr:0.001000 Backbone lr:0.001000 Mean:19.6 Std:8.5 Complete Fails:9
% SATNet lr:0.001000 Backbone lr:0.000100 Mean:90.4 Std:8.8 Complete Fails:1
% SATNet lr:0.001000 Backbone lr:0.000010 Mean:96.7 Std:0.2 Complete Fails:0
% SATNet lr:0.000100 Backbone lr:0.001000 Mean:17.0 Std:4.1 Complete Fails:8
% SATNet lr:0.000100 Backbone lr:0.000100 Mean:74.9 Std:8.8 Complete Fails:0
% SATNet lr:0.000100 Backbone lr:0.000010 Mean:96.5 Std:0.2 Complete Fails:0
% SATNet lr:0.000010 Backbone lr:0.001000 Mean:14.4 Std:3.4 Complete Fails:9
% SATNet lr:0.000010 Backbone lr:0.000100 Mean:31.8 Std:7.1 Complete Fails:5
% SATNet lr:0.000010 Backbone lr:0.000010 Mean:71.9 Std:5.4 Complete Fails:0

\subsection{Experiment 3}
The training accuracy rose from 96.7$\pm$0.2\%~(0) to 99.1$\pm$0.1\%~(0).
\clearpage
\subsection{Experiment 4}
\begin{table}[h]
  \caption{Effects of Different Neural Architectures on Training Accuracy}
  \label{table:architecture_train}
  \centering
  \begin{tabular}{ccccc}
    \toprule
    & & \multicolumn{2}{c}{Backbone Output Layer} \\
    \cmidrule(r){3-4}
    Architectures & Parameters & Softmax & Sigmoid \\
    \midrule
    LeNet \citep{lecun1998gradient} & 68,626 & 63.2$\pm$14.2\%~(4) & 99.1$\pm$0.0\%~(0)\\
    Sudoku CNN & 860,780 & 99.1$\pm$0.1\%~(0) & \textbf{99.5$\pm$0.0\%~(0)}\\ 
    ResNet18 \citep{he2016deep} & 11,723,722 & 67.6$\pm$6.2\%~(0) & 97.4$\pm$0.4\%~(0)\\
    \bottomrule
  \end{tabular}
\end{table}
% [63.214166666666664, 99.09700000000001, 67.5525, 99.05116666666666, 99.529, 97.36683333333332]
% [14.16874194004465, 0.05629496830792348, 6.228256633637091, 0.03290657369877016, 0.025166360066810178, 0.43225017404619204]
% [4, 0, 0, 0, 0, 0]
% [68626, 860780, 11723722, 68626, 860780, 11723722]

\subsection{Further Investigation into m and aux}
One of the reviewers proposed setting $m$ and $aux$ according to the relationship $m=out+aux$, where $out$ is the number of output variables. In the case of the MNIST mapping problem, we observed that while not necessarily optimal, it can be a good rule of thumb.

\begin{figure}[h]
\vspace{-0.1cm}
\centering
\includegraphics[width=\textwidth]{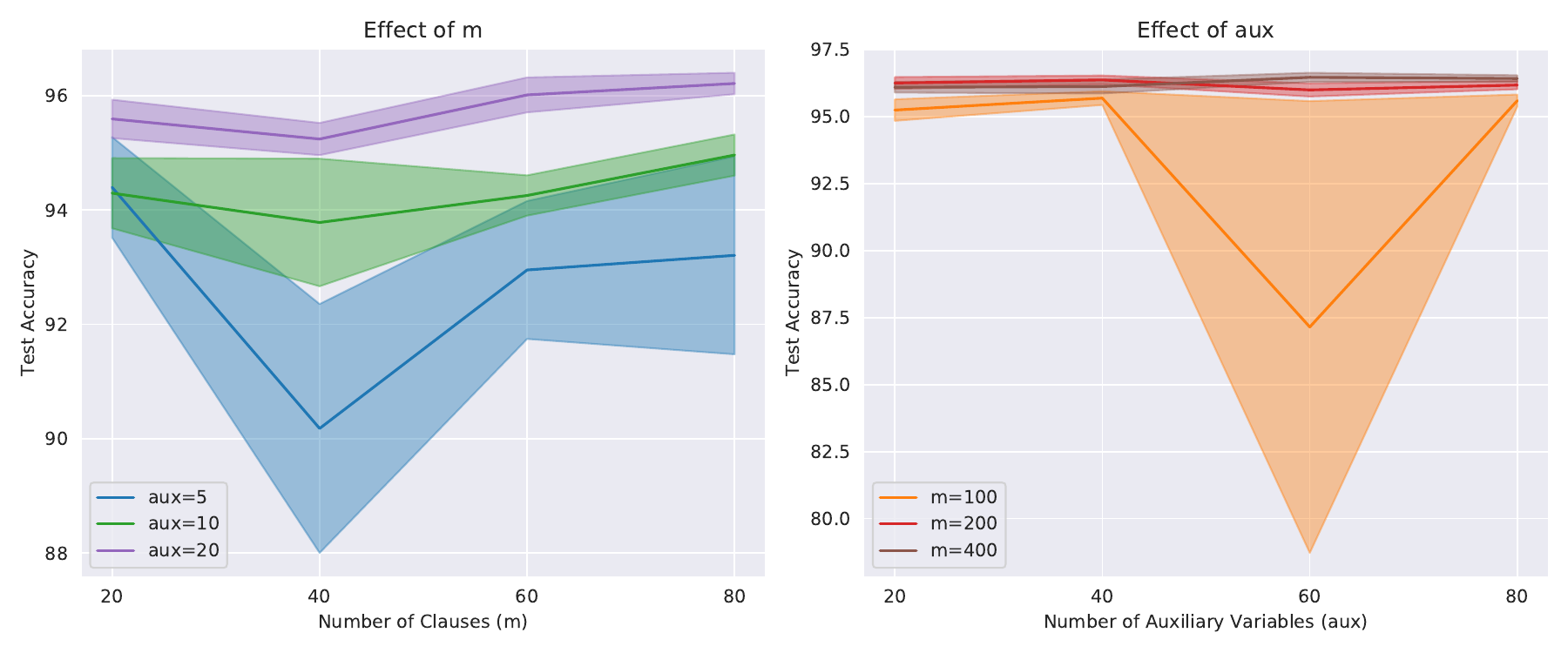}
\vspace{-0.5cm}
\caption{Both graphs show test accuracy on the MNIST mapping problem with the shaded interval representing the standard error.}
\label{fig:additional_m_aux}
% \vspace{-0.3cm}
\end{figure}

Another reviewer suggested that Experiment 1 be re-run with smaller values of $aux$. We show the results of re-running Experiment 1 with 10x smaller $aux$ in Figure \ref{fig:additional_m_aux}. We can observe that in this regime where $m$ is significantly higher than $aux$, larger $m$ and smaller $aux$ show a more muted benefit.

\end{document}